\pdfoutput=1
\documentclass{article}
\usepackage{ijcai27}
\usepackage{times}
\usepackage{url}
\usepackage[hidelinks]{hyperref}
\usepackage[utf8]{inputenc}
\usepackage[T1]{fontenc}
\usepackage[small]{caption}
\usepackage{graphicx}
\usepackage{amsmath}
\usepackage{booktabs}
\newcommand{\citep}[1]{\cite{#1}}
\newcommand{\citet}[1]{\citeauthor{#1}~\shortcite{#1}}
\title{Prompt Sensitivity in Tool-Augmented Model Comparisons}
\author{Mateusz Modrzejewski \and Tomasz Radzikowski
\affiliations
Institute of Computer Science, Warsaw University of Technology
\emails
\{mateusz.modrzejewski, tomasz.radzikowski\}@pw.edu.pl
}

\begin{document}

\maketitle

\begin{abstract}
Language models are compared on question-answering benchmarks that pose each question in one wording, and a reworded
question can change which of two models scores higher. We ask whether giving the models tools makes such comparisons
more stable. An accuracy gain cannot answer this, because tools change several steps between the question and the
scored answer at once. We propose a matched protocol that separates those steps: the answer readout, reasoning, the
decision to call, the tool's answer, and the report supplied without a call. We apply it to six open models in five
domains under five wordings. We find that tools raise accuracy where a program answers the question and widen the
wording sensitivity of many pairwise comparisons. The widening follows how often a model calls, and call frequency
follows the wording for some models and not others. It also depends on the interface and the instructions: a native
tool-calling interface and a second tool description leave no detected widening in molecules, and sensitivity persists
where tools return intermediate results. These findings show that deterministic reports do not guarantee stable
comparisons. We recommend reporting call frequency and comparison stability under several wordings beside accuracy.
\end{abstract}

\section{Introduction}
\label{sec:intro}

\begin{figure}[t]
\centering
\includegraphics[width=\columnwidth]{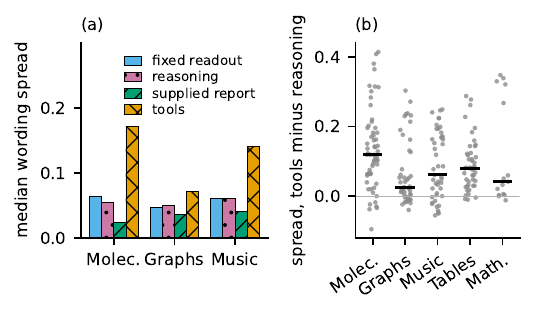}
\caption{(a) Median wording spread of a model pair by condition in the letter domains, six models. (b) Change in a
pair's spread when the tool loop replaces reasoning without tools, one point per pair and written form, median as a
bar.}
\label{fig:teaser}
\end{figure}

A benchmark asks a model a question in one wording and counts correct answers. The same question in another wording
gives another score. The difference between two models, which decides their order in a ranking, moves with the
wording by as much as it moves with a change of model \citep{sclar2023quantifying,mizrahi2023state}. Multi-prompt
evaluation proposes to report several wordings. Tool use raises a further question. In a tool loop of the kind ReAct
introduced \citep{yao2022react}, the model writes a call, a program answers, and the model reads the answer before it
decides. The program's answer is the same under every wording, but nothing guarantees that the model makes the same
call under every wording.

One molecule item shows the difference. It asks for the number of heavy atoms in the molecule
\texttt{C=CC1=C(C=CC(=C1)Br)Br}, with four options. Given a heavy-atom tool, Qwen3-8B calls it under four of our five
wordings and answers correctly each time. Under the remaining wording, which shows the molecule before the question,
it counts the atoms in prose, makes no call before its generation budget ends, and the letter read at the appended
answer line is wrong. When the tool's report is printed in the prompt, the model answers correctly under all five
wordings. The report is the same in every case. What differs is whether the model obtains it.

An accuracy gain under tools cannot show whether tools stabilize comparisons, because several steps change at once.
The answer is no longer read from the probability of the first option letter, a common readout that for
instruction-tuned models often disagrees with the letter the model writes \citep{wang2024look}. The model writes text
before it answers. It decides whether to call a tool after reading the wording. It receives the tools' reports. Any of
these steps can add or remove wording sensitivity. A comparison of the same model with and without tools mixes them.

We separate the steps on the same items with matched conditions: the first-token read, an answer read at an answer
line, reasoning before that line, the tools' reports supplied in the prompt, a plain-text tool loop over deterministic
programs, the same loop with tools that answer ``no information'', and for three models the native tool-calling
interface. Every condition runs under five wordings of the instruction and under each written form of the input. For each pair of models we measure the wording spread, the range of
their accuracy difference across the wordings. The tools' information and the model's call behavior can affect this
spread in different ways. Supplying the report under every wording measures the comparison with direct access to the
information. Letting the model decide whether to obtain it exposes the comparison to changes in calling as well. We
compare the two and intervene on the interface, the instructions, the budget and the reports' reliability to see when
the spread changes.

Tools raise accuracy where a program answers the question and still widen the wording spread of many model pairs
against reasoning without tools (Figure~\ref{fig:teaser}). In molecules the median spread of a pair is 0.17 with the
plain-text loop and 0.02 with the reports supplied. There, a model that calls is almost always right, so its accuracy
follows its call frequency, and for several models the call frequency follows the wording. The widening depends on
the interface and the instructions. In molecules, the native interface and a second tool description leave no
detected widening in the model groups tested. Forcing one call and removing the wordings' output clause help some
models and not others. Wrong reports shrink the widening while the call frequencies stay. Sensitivity persists under
the native interface where the tools return intermediate results.
Reading the answer from the first token adds sensitivity of its own on graphs, which an answer line removes.

\paragraph{Scope.} We study open models of 4B to 72B parameters with greedy decoding, a plain-text tool loop and, for
three model families, the native interface, wordings in English, and deterministic programs that read a canonical form
of the item. The quantity is the stability of a comparison between two models across wordings rather than the
robustness of one model, and the spreads are statistics of our wording set. The molecule and graph tasks are built so
that an operator answers the question, which isolates the call and the use of the report rather than tool discovery
or argument construction. The question is how far the comparisons that tools produce can be trusted. The results
say which evaluation choices change that.

This paper makes three contributions. First, we propose a matched protocol that measures how the answer readout,
reasoning, the decision to call, the tool interface and access to the tools' reports change the stability of pairwise
model comparisons across wordings. Second, we split each model's accuracy changes across wordings into a
call-frequency component and a conditional-accuracy component and relate them to the movement of pairwise differences.
Third, we apply both to six open models in five domains, a Qwen3 size series and a 72B model, and test the result
against seven interventions on the call. Paraphrases of the user's request under tool calling
\citep{rabinovich2025robustness,zhou2026simulation,chen2026lost} and the decision to call
\citep{ross2025when2call,wu2026call,cheng2026knowing,sun2026when2tool} are established objects of study. Our focus is
the stability of pairwise comparisons of executed answers, read through matched changes to the readout, the call and
access to the reports. The wordings, the tool descriptions and the supporting tables are in the supplementary
material (Sections S1 to S4).

\section{Evaluation Conditions and Measures}
\label{sec:method}

\paragraph{Pairwise wording spread.} For two models $a$ and $b$ and a set of wordings $W$, let $\mathrm{acc}_a(w)$ be
the accuracy of model $a$ under wording $w$. The wording spread of the pair is
\begin{equation}
  S_{ab} = \max_{w \in W} D_{ab}(w) - \min_{w \in W} D_{ab}(w),
\end{equation}
where $D_{ab}(w) = \mathrm{acc}_a(w) - \mathrm{acc}_b(w)$ is the difference that orders the two models, so the spread
is its range over wordings. We compute it for each condition and each written
form of the input, on the items present for both models under every wording, where a cell enters only when it holds at
least 0.9 of the domain's items and an answer that cannot be parsed counts as incorrect. Intervals come from a paired
bootstrap over items. We also report each model's own accuracy range across wordings.

\paragraph{The conditions.} All conditions use the same items, wordings and written forms, and decoding is greedy.
Under the \emph{first-token read}, the answer to a four-option item is the option letter with the highest probability
among the four letters at the first position of the response. Under the \emph{fixed readout}, an answer line (the text
``ANSWER:'') follows the prompt, and the letter is read there with the same restricted read. For a numeric item the
model writes the number on that line, and we call this the direct answer. At the read position we record the letter
mass, the share of the full next-token probability that falls on the four letters, after any formatting mark such as a
markdown bold mark. Under \emph{reasoning}, the model writes freely before the answer line. Under \emph{tools}, a fixed
tool description precedes the unchanged wording. The model may write call lines, each naming one tool, each tool's
report is appended as a line, and the episode ends at an answer line, after six calls, or at the token budget, where an
answer line is appended and read. Under \emph{constant tools}, the loop and the description are unchanged and every
tool answers ``no information''. Under the \emph{supplied report}, the report that the loop would print for the item's
question type is appended to the fixed readout's prompt, with no tool description, no loop and no call. Under the
\emph{forced call}, the loop refuses an answer without a call once. Under the \emph{native interface}, the same tools
are declared through the model's own tool-calling format, so a call is a structured turn rather than a text line.

\paragraph{Interpretation of the controls.} The conditions are matched diagnostic contrasts rather than a partition of
causes, and each has a reading fixed before its result. From the first-token read to the fixed readout, the position
and the context of the read change. Reasoning adds the model's own text before the answer. The supplied report adds the
tool's information at the start of the prompt and nothing the model must do to obtain it. The tool loop adds the
description, the call syntax, the reports and the loop's limits together. Constant tools keep that interface and
remove the information, which can also change how often the model calls again and stops. A difference between tools
and constant tools is read as the information's and a difference between constant tools and reasoning as the loop's. Chance, 0.25 on a four-option item, and the supplied report are the two reference values, neither a bound on
what a model can reach. If the spread under tools lies close to the supplied report's, the information settles the
comparison. If it exceeds both the supplied report's and the fixed readout's, follows the call frequency where a tool
answers the question, and falls under an intervention that makes the model call under every wording, the call
behavior carries it.

\paragraph{Call-frequency decomposition.} In a tool condition, a model's accuracy under wording $w$ is an identity in
its call frequency $c_w$, the share of episodes with at least one call, and the accuracies $a^1_w$ and $a^0_w$ of the
episodes with and without a call:
\begin{equation}
  \mathrm{acc}(w) = a^0_w + c_w\,(a^1_w - a^0_w) .
  \label{eq:identity}
\end{equation}
The difference $a^1_w - a^0_w$ is observed on different episodes and is not the causal effect of a call. With a
reference $(c_*, a^1_*, a^0_*)$ pooled over the five wordings' episodes, the change $\mathrm{acc}(w) - \mathrm{acc}_*$
splits exactly into a call-frequency component and a conditional-accuracy component by the two-factor Shapley rule:
\begin{equation}
  C_w = \tfrac{1}{2}(c_w - c_*)\big[(a^1_w - a^0_w) + (a^1_* - a^0_*)\big],
  \label{eq:callterm}
\end{equation}
and the rest is the conditional component. A wording without any episode in one stratum takes the reference value
there. For a pair of models, the call share is the covariance over wordings of the pair difference with the
difference of the two models' $C_w$, divided by the variance of the pair difference, with forms pooled. The two
components' shares sum to one, a share can lie outside $[0, 1]$ when the components move against each other, and
the share describes covariation with the pair difference rather than a decomposition of the range in Equation 1.
The call frequency counts whether any tool was called and not which items or which tools, so a constant frequency
does not establish that the same items receive calls.

\section{Experimental Setup}
\label{sec:setup}

\begin{table}[t]
\centering
\scriptsize
\setlength{\tabcolsep}{2.5pt}
\begin{tabular}{@{}lp{2.35cm}p{3.6cm}@{}}
\toprule
Domain & Items & Written forms; what the tools return \\
\midrule
Molecules & 504, four options, PubChem & SMILES, random SMILES, InChI, IUPAC name; RDKit answers every question type \\
Graphs & 504, four options, generated & edge list, neighbor list, friendship; NetworkX answers every type \\
Music & 361, four options, ABC-Eval & ABC, MusicXML, note events; music21 answers bar count, and key and time partly \\
Tables & 500, numeric, FinQA & markdown, CSV, sentences; rows, columns and a calculator \\
Mathematics & 336, numeric, MATH-500 & one form; a calculator and an equation solver \\
\bottomrule
\end{tabular}
\caption{The domains, written forms and tools. Every written form of an item shares one canonical tool input and gold
answer. In molecules and graphs an operator answers the question directly; in tables and mathematics the tools return
values the model must combine.}
\label{tab:domains}
\end{table}

\paragraph{Domains.} The five domains differ in what the tools answer (Table~\ref{tab:domains}). Every tool reads one
canonical form of the item, so its reports are identical across written forms and wordings. The molecule items ask for
six RDKit properties of compounds drawn from PubChem\footnote{\url{https://pubchem.ncbi.nlm.nih.gov}}, 84 per property.
The graph items follow the tasks of \citet{fatemi2023talk} on random graphs of 5 to 20 nodes, 84 per task. The gold
answers of both domains come from the same libraries as the tools, so these are controlled tasks with direct-answer
operators. The music items are those of ABC-Eval \citep{zhao2025abc} that fit a common token budget in three
notations, with music21 operators \citep{cuthbert2010music21}. The table items are FinQA questions answerable from the
table alone \citep{chen2021finqa}, and the mathematics items the MATH-500 problems with a plain numeric answer
\citep{hendrycks2021measuring,lightman2023verify}. A domain entered only if, on its first 200 items under the fixed
readout, Qwen3-8B or Phi-4 scored at most 0.90 and their pair spread reached 0.02, which excluded GSM8K
\citep{cobbe2021training}, where Qwen3-8B and Phi-4 score 0.95 and 0.97. Music, where the protocol was developed,
was exempt from the rule. The wordings do not mention tools. A letter answer is correct when it is the gold option. A
mathematics answer is correct when it equals the gold number. A table answer is correct when it lies within 1\% of
the executed answer or within 0.005 of it, or when it has at least one decimal and equals the answer rounded to the
decimals the model wrote (two decimals when the answer is below one). For a question whose gold value is a ratio, the
same rules apply to the value times 100. Integers never match through rounding.

\paragraph{Models and harness.} Six open instruction-tuned models, each at a fixed revision: Qwen3-8B with its thinking
mode off, Phi-4, Llama-3.1-8B-Instruct, Gemma-3-12B, Mistral-Small-24B and Qwen3-30B-A3B, a mixture-of-experts model
\citep{yang2025qwen3,abdin2024phi4,grattafiori2024llama3,gemma2025gemma3,mistral2025small}. Three of them run with
8-bit weights. Qwen3-4B, Qwen3-14B and Qwen3-32B, dense models, form a size series with the two Qwen3 models of the
grid, and Qwen2.5-72B-Instruct, a model of the previous generation with 8-bit weights, is the largest, both on
molecules, graphs and music. The loop generates 64 tokens per step in the letter domains and 256 in the number
domains, within an episode budget of 256 tokens in music and molecules, 384 on graphs, 1,024 in tables and 1,536 in
mathematics. The models' outputs are the data of this study: every answer, call line and trajectory analysed here was
generated by these models under the conditions described, with greedy decoding, on vLLM. The revisions, budgets, tool
descriptions and the native schemas are in the supplementary material (Section S2).

\paragraph{Analysis.} Spreads and accuracies carry 95\% percentile intervals from 1,000 bootstrap resamples over items.
A contrast of two conditions is the difference of a pair's two spreads under one resample of the items, its two-sided
bootstrap p-value is twice the share of resamples on the other side of zero, and Holm correction runs across the model
pairs of one domain, written form and contrast. We count a pair as wider or narrower when the corrected contrast
excludes zero, and we report the raw differences beside the counts. Under a simulated null of equal, nonzero wording
effects, the bootstrap rejects single comparisons at the 0.05 level in 0.12 of 600 simulated contrasts for tools
against the fixed readout in molecules (0.035 after Holm) and in at most 0.04 in the other contrasts checked. For that
one row we also report a within-item exchange test, which permutes the two conditions' outcome vectors within each item
and rejects in 0.07 of 180 simulated contrasts, 0 after Holm (the null's construction and both procedures' counts for
every contrast are in Section S3). Because neither procedure is shown to hold the nominal rate when the two conditions differ in accuracy level, the
counts are exploratory flags, and conclusions about size rest on the raw paired differences of spread and on the
interventions. Spreads below 0.05 are treated as zero when we
read them, and every contrast is tested on the raw spreads. Repeating a full cell under batched greedy decoding changes
the predictions of at most 0.07 of the rows in three letter-domain cells and 0.49 in one table cell, and moves any
pair's spread by at most 0.019 and 0.036. Mathematics has no full-size repeat. The smallest detected widening against
reasoning is 0.09 in molecules and 0.10 in tables, above those moves.

\section{Results}
\label{sec:results}

\begin{table}[t]
\centering
\scriptsize
\setlength{\tabcolsep}{1.4pt}
\begin{tabular}{@{}lcccccc@{}}
\toprule
 & Qwen3 8B & Phi-4 & Llama 8B & Gemma 12B & Mistral 24B & Qwen3 30B \\
\midrule
\multicolumn{7}{@{}l}{\textit{Molecules}} \\
\quad first-token read & 0.36 & 0.46 & 0.34 & 0.41 & 0.47 & 0.46 \\
\quad fixed readout & 0.43 & 0.52 & 0.38 & 0.42 & 0.50 & 0.49 \\
\quad reasoning & 0.52 & 0.57 & 0.46 & 0.54 & 0.55 & 0.55 \\
\quad supplied report & 0.99 & 0.98 & 0.99 & 0.98 & 1.00 & 1.00 \\
\quad tools & 0.92 & 0.75 & 0.94 & 0.98 & 0.94 & 1.00 \\
\quad constant tools & 0.44 & 0.51 & 0.33 & 0.42 & 0.45 & 0.49 \\
\midrule
\multicolumn{7}{@{}l}{\textit{Graphs}} \\
\quad first-token read & 0.34 & 0.71 & 0.57 & 0.75 & 0.78 & 0.71 \\
\quad fixed readout & 0.73 & 0.80 & 0.63 & 0.79 & 0.84 & 0.81 \\
\quad reasoning & 0.90 & 0.90 & 0.75 & 0.91 & 0.90 & 0.92 \\
\quad supplied report & 0.97 & 0.96 & 0.93 & 0.97 & 0.98 & 0.98 \\
\quad tools & 0.97 & 0.92 & 0.83 & 0.97 & 0.96 & 0.96 \\
\quad constant tools & 0.81 & 0.83 & 0.53 & 0.73 & 0.76 & 0.83 \\
\midrule
\multicolumn{7}{@{}l}{\textit{Music}} \\
\quad first-token read & 0.39 & 0.41 & 0.38 & 0.39 & 0.41 & 0.43 \\
\quad fixed readout & 0.40 & 0.43 & 0.42 & 0.39 & 0.44 & 0.45 \\
\quad reasoning & 0.41 & 0.43 & 0.38 & 0.42 & 0.47 & 0.46 \\
\quad supplied report & 0.59 & 0.59 & 0.59 & 0.57 & 0.60 & 0.59 \\
\quad tools & 0.51 & 0.54 & 0.54 & 0.54 & 0.56 & 0.60 \\
\quad constant tools & 0.40 & 0.44 & 0.38 & 0.38 & 0.42 & 0.48 \\
\midrule
\multicolumn{7}{@{}l}{\textit{Tables}} \\
\quad direct answer & 0.43 & 0.54 & 0.39 & 0.46 & 0.63 & 0.61 \\
\quad reasoning & 0.73 & 0.75 & 0.55 & 0.71 & 0.74 & 0.74 \\
\quad tools & 0.62 & 0.57 & 0.22 & 0.61 & 0.46 & 0.70 \\
\quad constant tools & 0.54 & 0.53 & 0.19 & 0.29 & 0.45 & 0.68 \\
\midrule
\multicolumn{7}{@{}l}{\textit{Mathematics}} \\
\quad direct answer & 0.33 & 0.33 & 0.15 & 0.22 & 0.27 & 0.46 \\
\quad reasoning & 0.82 & 0.84 & 0.56 & 0.85 & 0.77 & 0.80 \\
\quad tools & 0.82 & 0.76 & 0.30 & 0.86 & 0.68 & 0.64 \\
\quad constant tools & 0.79 & 0.76 & 0.20 & 0.84 & 0.65 & 0.65 \\
\bottomrule
\end{tabular}

\caption{Mean accuracy by condition and domain, with equal weight over wordings and written forms; an unparsed answer
is incorrect. In tables and mathematics the fixed readout is the direct answer and the supplied report is not defined.}
\label{tab:acc}
\end{table}

\begin{table}[t]
\centering
\scriptsize
\setlength{\tabcolsep}{2.6pt}
\begin{tabular}{@{}lccccc@{}}
\toprule
 & Molec. & Graphs & Music & Tables & Math. \\
\midrule
pairs $\times$ written forms & 60 & 45 & 45 & 45 & 15 \\
\midrule
fixed readout vs.\ first-token read & 0/5 & 0/36 & 0/3 & -- & -- \\
reasoning vs.\ fixed readout & 0/0 & 0/1 & 0/0 & 0/0 & 0/0 \\
supplied report vs.\ fixed readout & 0/21 & 0/5 & 0/0 & -- & -- \\
tools vs.\ fixed readout & 39/1$^{\dagger}$ & 10/0 & 19/0 & -- & -- \\
tools vs.\ supplied report & 54/0 & 16/0 & 22/0 & -- & -- \\
tools vs.\ reasoning & 37/2 & 12/0 & 17/0 & 15/0 & 5/0 \\
constant tools vs.\ reasoning & 1/0 & 10/0 & 0/0 & 32/0 & 5/0 \\
\midrule
call share under tools & 0.94 & 0.75 & 0.57 & 0.06 & 0.07 \\
\bottomrule
\end{tabular}

\caption{Pairs and written forms flagged wider/narrower under the first condition than the second, after Holm
correction within domain, form and contrast, six models; the flags are exploratory (Section~\ref{sec:setup}) and a zero
does not establish equivalence. $\dagger$: the exchange test (bootstrap 34/2). Last row: the call share.}
\label{tab:contrasts}
\end{table}

\paragraph{Repeatability and implementation checks.} A
separate run of the first-token read on the music items, with other hardware and code, agrees with ours on 0.95 to
0.99 of the answers of each of five models. Two runs of 200 items at batch size one give identical answers. Run alone,
the tools return the gold answer by construction in molecules and graphs. In music they return it on 0.99 of the
bar-count items, 0.63 of the key items and 0.53 of the time-signature items. Three models run again in bfloat16 move
accuracy by at most 0.022 in 59 of 60 cells and by 0.06 in one. No contrast changes direction.

\paragraph{Accuracy gains and tool coverage.} In molecules tools raise accuracy over the fixed readout by 0.23 to
0.56, on graphs by 0.11 to 0.24 and in music by 0.10 to 0.15, for every model (Table~\ref{tab:acc}). Constant tools
move it by at most 0.10 in either direction, so the gain comes with the tools' reports rather than with the loop
alone. The gain follows what the operators know. In music it appears on bar count, where the operator is right on 0.99
of the items, and turns into a loss of 0.04 to 0.10 on key, where the operator is right on 0.63. In tables and
mathematics the tools return values that the model must still combine, and accuracy falls below reasoning without
tools for most models. Constant tools lower it by similar amounts, so there the loop costs more than the tools give.
On the same graph items with only the two operators that return a node's neighbors and whether two nodes share an
edge, the gain over the fixed readout goes for Llama-3.1-8B and Phi-4, and constant tools widen where the tools no
longer do (Section~\ref{sec:mechanism}). A practitioner gains from tools on the question types that a program answers
better than the model.

\paragraph{Answer-prefix scoring on graphs.} Reading the answer at an answer line instead of the first token narrows
the spread of 36 of 45 graph pairs, and the median falls from 0.25 to 0.05 (Table~\ref{tab:contrasts}). Qwen3-8B places a median mass of 0.00 on the letters at the first token on graphs, but
Gemma-3-12B and Qwen3-30B-A3B place 1.00 there and still lose 0.21 and 0.33 of their median graph spread. Over the 150
pair-by-form rows of the letter domains, the reduction correlates with the accuracy that the first-token read hides
(Spearman 0.58) and not with the letter mass ($-0.12$). In molecules and music the readout changes almost nothing,
permuting the options under the fixed readout changes the Holm outcome of no row, and reasoning before the answer
line widens no pair in any domain. Multiple-choice scores taken from the first token are therefore safer read at an
answer line.

\paragraph{Pairwise sensitivity under tool use.} Against reasoning without tools, the evaluator's baseline for an
agent, the plain-text loop widens the spread of 37 of 60 molecule pairs and of a quarter to two fifths of the pairs in
each other domain, and narrows two pairs in all (Table~\ref{tab:contrasts}). The median change of a pair's spread is
0.12 in molecules, 0.08 in tables, 0.06 in music, 0.04 in mathematics and 0.03 on graphs, and 0.73 to 0.93 of the
pairs move upward (Figure~\ref{fig:teaser}b). Against the fixed readout the molecule count is 39 of 60 under the
exchange test. The rows share models and items and are not independent detections. The widening is concentrated: on graphs every widened pair includes Llama-3.1-8B, and in
mathematics every one includes Qwen3-30B-A3B. Under the supplied report no model's own accuracy range across wordings
exceeds 0.06 in any letter domain, and under tools it reaches 0.24 for Llama-3.1-8B on graphs and 0.36 for
Qwen3-30B-A3B in mathematics, so the pairs that widen are the pairs that contain such a model. The sign changes that
leaving one wording out produces concentrate on the wording of each domain that lowers call frequencies most, and with
ten wordings the range under tools grows over its five-wording value by a median factor of 2.88 in molecules, against
1.12 under the fixed readout.

\paragraph{Ranking reversals and selection loss.} A pair's order reverses when its accuracy difference exceeds 0.05
under one wording and falls below $-0.05$ under another. Under tools 4 of 15 pairs reverse in music and in molecules and
2 in mathematics, against none under the fixed readout or under reasoning in any domain. One wording gives the same
order of a pair as the mean over all five for 0.76 of the pairs in music and 0.85 in molecules under tools, against
about 0.9 under the fixed readout. On five further wordings written for three models, 2 of the 3 molecule pairs
reverse under tools and none under the fixed readout. Choosing the best model under one wording and scoring it under the other
four loses 0.03 of accuracy on average in music under tools and less than 0.01 elsewhere. A leaderboard of tool-using models built on one wording orders some pairs
differently depending on the wording chosen, at a small average cost to the model it selects.

\section{Tool Invocation and Interface Effects}
\label{sec:mechanism}

\begin{figure*}[t]
\centering
\includegraphics[width=\textwidth]{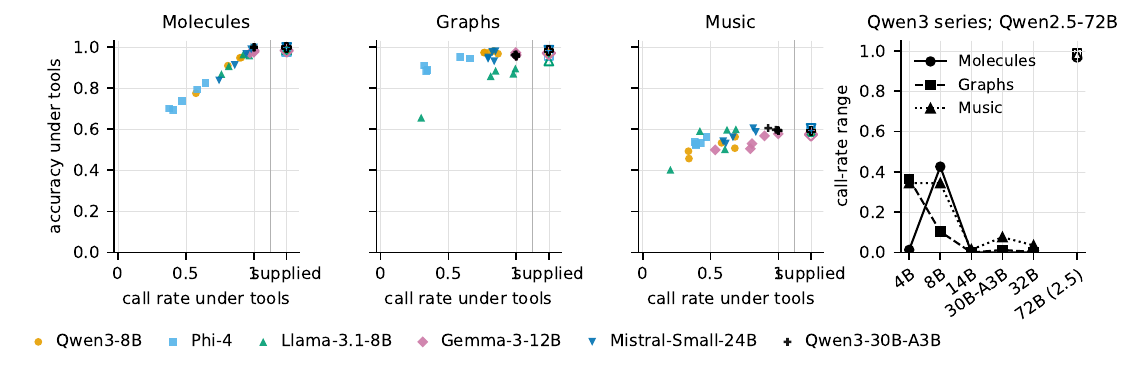}
\caption{Left: accuracy under tools against call frequency, one point per model and wording, written forms pooled;
open markers are the accuracy with the report supplied. Right: the call-frequency range of five Qwen3 models (30B-A3B
is a mixture of experts) and, as an open marker, Qwen2.5-72B of the previous generation; the panel does not isolate
size.}
\label{fig:decision}
\end{figure*}

\begin{table}[t]
\centering
\scriptsize
\setlength{\tabcolsep}{3pt}
\begin{tabular}{@{}lcccc@{}}
\toprule
 & \multicolumn{3}{c}{call-rate range (mean call rate)} & pairs wider \\
\cmidrule(lr){2-4}
Condition & Qwen3 8B & Phi-4 & Llama 8B & than fixed \\
\midrule
plain-text loop (reference) & 0.43 (0.84) & 0.27 (0.50) & 0.22 (0.89) & 10 of 12 \\
episode budget $\times 4$ & 0.41 (0.85) & 0.25 (0.53) & -- & 4 of 4 \\
10\% of reports wrong & 0.43 (0.83) & 0.22 (0.53) & 0.23 (0.88) & 8 of 12 \\
30\% of reports wrong & 0.42 (0.84) & 0.27 (0.50) & 0.23 (0.89) & 6 of 12 \\
50\% of reports wrong & 0.42 (0.84) & 0.25 (0.49) & 0.23 (0.89) & 0 of 12 \\
output clause removed & 0.35 (0.83) & 0.21 (0.68) & 0.08 (0.95) & 7 of 12 \\
third tool description & 0.34 (0.88) & 0.30 (0.64) & 0.11 (0.98) & 9 of 12 \\
no-call answer refused once & 0.01 (0.99) & 0.14 (0.77) & 0.21 (0.91) & 1 of 12 \\
second tool description & 0.07 (0.97) & 0.06 (0.94) & 0.00 (1.00) & 0 of 12 \\
\midrule
 & Qwen3 8B & Llama 8B & Mistral 24B & \\
\midrule
plain-text loop (reference) & 0.43 (0.84) & 0.22 (0.89) & 0.26 (0.90) & 10 of 12 \\
native interface & 0.00 (1.00) & 0.00 (1.00) & 0.00 (1.00) & 0 of 12 \\
\midrule
supplied report, six models & \multicolumn{3}{c}{no call} & 0 of 60 \\
\bottomrule
\end{tabular}

\caption{Interventions on the call in molecules: each model's call-frequency range across wordings (mean in
parentheses) and the pairs flagged wider than under the fixed readout (exploratory, as in Table~\ref{tab:contrasts}).
Upper block: Qwen3-8B, Phi-4 and Llama-3.1-8B against their plain-text loop; the budget row covers the Qwen3-8B and
Phi-4 pairs. Lower block: the native interface for Qwen3-8B, Llama-3.1-8B and Mistral-Small-24B against the same
models' plain-text loop. A range of 0.00 is a constant frequency; the mean says whether it is one.}
\label{tab:interventions}
\end{table}

Deterministic reports do not guarantee stable comparisons when the model decides whether to obtain them. With the
plain-text loop, the median spread in the letter domains exceeds both the supplied report's and the fixed readout's
(Figure~\ref{fig:teaser}a), and tools are wider than the supplied report in 54 of 60 molecule pairs
(Table~\ref{tab:contrasts}). This section relates the widening to call frequency. It then reads the supplied-report controls and seven
interventions on the call before turning to the domains where the tools return intermediate results.

\paragraph{Call frequency and conditional accuracy.} In molecules the call share is 0.94 [0.88, 1.02], on graphs 0.75
and in music 0.57, against under 0.1 in tables and mathematics (Table~\ref{tab:contrasts}; intervals in Section S4).
Figure~\ref{fig:decision} shows why. In molecules a model that calls is right in 0.97 to 1.00 of episodes under every
wording, so every model's accuracy lies close to one line rising with its call frequency, consistent with
Equation~\ref{eq:identity}. What the identity does not fix is which models move. Gemma-3-12B and
Qwen3-30B-A3B call under every wording on graphs, while Llama-3.1-8B calls in nearly every episode under one wording and
in fewer than a third under another (the call frequency of every model under every wording is Figure~S1 of the
supplementary material). A wording that lowers a model's call frequency costs
most where the operator is best.

\paragraph{Supplied-report controls.} With the report supplied at the start of the prompt and no call to make, no
pair's spread is wider than under the fixed readout in any of the 150 letter-domain rows, and every model is within
0.01 of its highest accuracy (Figure~\ref{fig:decision}, open markers). The stability does not depend on the ceiling
or on the absence of reasoning. When the model reasons before answering with the report in its prompt, its spread is
wider than the fixed readout's in 0 of the 21 rows of three models (three pairs over the seven graph and molecule
forms). When 20\% of the graph reports and 27\% of the molecule reports are replaced by a wrong option, accuracy
falls by up to 0.26, and the spread is still wider than the fixed readout's in 0 of the 105 rows of six models and
narrower in 29. In music the supplied report reaches 0.57 to 0.60, far from the ceiling, and never widens. We
read the supplied report as a diagnostic of report access, not of what a model knows.

\paragraph{Interface, instruction and report-quality interventions.} Table~\ref{tab:interventions} reads seven
interventions on the molecule loop by the call-frequency range and the number of widened pairs. The native interface
and the second tool description bring every model of their group to a mean call frequency of 0.94 or more with a range
of 0.07 or less, and leave no widened pair where the plain-text loop widens 10 of 12 in each group. On graphs the
native interface reduces the widened pairs from 4 to 1 of 9 and the call share from 0.75 to 0.25. Refusing a no-call
answer once brings Qwen3-8B to a range of 0.01 and the count to 1 of 12, while Phi-4's range falls from 0.27 to 0.14
and Llama-3.1-8B's stays. Removing the output clause takes Llama-3.1-8B's range from 0.22 to 0.08 and leaves Qwen3-8B's
and Phi-4's above 0.2, with 7 of 12 pairs wider. A third description leaves every range above 0.1 and 9 pairs wider, so
the widening holds under two of the three descriptions and vanishes under the one that makes every model call. Two
interventions change the loop without changing the call. A four-fold token budget lowers the share of episodes cut at
the budget from 0.12 and 0.22 to 0.01 and 0.02 for Qwen3-8B and Phi-4 and leaves their ranges in place, so for these
two models the missed calls are not truncation. Replacing 10, 30 and 50\% of the reports by a wrong option after the
call leaves every range in place and takes the widening down step by step to none, which is what
Equation~\ref{eq:identity} gives when the reports' usefulness after a call shrinks. Sampling at temperature 0.7 keeps Qwen3-8B's ordering of the wordings by call frequency and its range. The interventions
therefore have model- and group-specific effects.

\paragraph{Model-specific instruction effects.} The wordings differ in their output-format clause, and in the letter
domains the one that asks for the answer alone lowers call frequencies most. For Llama-3.1-8B the clause accounts for its graph behavior:
without it, its graph call-frequency range falls from 0.69 to 0.09 and the widened graph pairs from 4 to 0 of 9. The
same holds for Qwen2.5-72B, which calls in at most 0.03 of episodes under that wording and in at least 0.61 under the
others, so that its tool spreads lie 0.34 to 0.48 above the fixed readout's in molecules, and which calls under every
wording once the clause is removed. For Qwen3-8B and Phi-4 the clause is a small part. Their ranges fall by less than
0.1, and 7 of 12 molecule pairs stay wider. Their calling responds to the paraphrase crossed with the tool description:
moving the description into the system turn changes a model's per-wording call frequency by up to a half, and the IUPAC name in place of the SMILES lowers Qwen3-8B's call
frequency by a third. Ranked by the call frequency they produce, the five wordings barely agree across the six models
(Kendall's W at most 0.13), so no single wording is the bad one for every model. Size alone does not order the
models either. In the Qwen3 series the 4B and 8B models change their call frequency with the wording and the models
from 14B call under almost every wording, while the 72B model of the previous generation has the widest range we
measured (Figure~\ref{fig:decision}, right). The series varies architecture as well as size.

\paragraph{Intermediate-result tools and stopping behavior.} In tables and mathematics the call share is under 0.1.
The share is a covariance ratio: the rest of the variation can lie in which items receive calls, which tool and
arguments are chosen, and what the model does with the values. Constant tools widen 32 of 45 table pairs, more than
working tools (Table~\ref{tab:contrasts}). Under constant tools four models keep calling the empty tools until the call
budget ends the episode, in 0.17 to 0.68 of table episodes, and removing the episodes that end on the budget or contain
a rejected call, a descriptive subset, leaves about half of the table widening and all of the mathematics widening.
The native interface cuts Llama-3.1-8B's budget stops from 0.61 to 0.08 of table episodes, yet Qwen3-8B's native
constant control still ends on the budget in 0.44 of table episodes and 0.37 of mathematics episodes, so the stopping failure persists under the
tested native interface. Under that
interface the three-pair table widening falls from 6 to 4 of 9, every remaining pair contains Mistral-Small-24B,
whose native call frequency in tables ranges from 0.28 to 0.90 across wordings, and in mathematics Qwen3-8B's native
call frequency ranges from 0.24 to 0.84. In the groups tested, native calling therefore removes the detected molecule widening and
reduces the graph widening from 4 to 1 of 9 pairs. Where the tools return intermediate results, its effects are
model-specific.

\section{Evaluation Recommendations}
\label{sec:recommendations}

The findings translate into four diagnostics for authors of benchmarks and leaderboards that compare models with and
without tools.
\begin{enumerate}
\item Check the answer-scoring protocol. For multiple-choice items, compare first-token scoring with scoring at an
answer line and report the difference. The letter mass alone does not validate the read.
\item Evaluate the same model comparisons under several wordings, and report accuracy, pairwise differences and call
frequency together. A changing call frequency is a diagnostic to read beside the comparison, and aggregate frequency
is not agreement on which items receive calls.
\item Run the supplied report under the same wordings. A model that is stable with the reports supplied and unstable
with tools differs from its rivals in obtaining the reports, which says nothing by itself about what it knows.
\item Report stops at the token or call budget, rejected and repeated calls, and the interface used. These matter
most where the tools return intermediate results.
\end{enumerate}
For developers of tool-using models, the target is calling that is consistent and appropriate under equivalent
instructions. A constant call frequency alone does not show it.

\section{Limitations}
\label{sec:limitations}

The spreads are ranges over five wordings, a statistic of this wording set that grows with ten wordings, and five
wordings support little inference to new ones. The intervals reflect the sampling of items only. Three models run with 8-bit weights. Mathematics has no full-size repeat, and in tables about half of a model's
predictions change between repeats while pair spreads move by at most 0.036. The tools read a canonical form of the
item and, in molecules and graphs, answer the question directly, so the gains are bound to the operators we wrote and
the tasks say little about tool discovery or argument construction. Domains entered on their fixed-readout spread and music was exempt, so the five domains are not a
sample of tasks and the prevalence of the effect outside them is unknown. The description variants and the native
interface were run for three models only, so the plain-text loop remains the protocol of the six-model tables. The call-frequency components are an algebraic attribution of
observed accuracy changes, not causal effects of calling. The wordings are in English. With its thinking mode on, Qwen3-8B
reasons instead of calling in most episodes, which we leave to future work. The supplied report is a
control rather than an agent. The forced call moves the calling fully for one model only.

\section{Related Work}
\label{sec:related}

\paragraph{Prompt sensitivity and the readout.} Accuracy varies by up to 76 points across equivalent prompt formats in
the setting of \citet{sclar2023quantifying}, and model rankings change across instruction paraphrases in
\citet{mizrahi2023state}, who call for evaluation over several prompts, and \citet{polo2024efficient} estimate the
distribution over many templates from a few evaluations. \citet{wang2024look} report that the
first-token probability over option letters often does not match the text answer of instruction-tuned models, and
\citet{alzahrani2024when} and \citet{zheng2023large} show that option order and answer selection move leaderboard
rankings. \citet{hua2025flaw} find that heuristic answer extraction makes much of the reported prompt sensitivity and
that an LLM judge restores the agreement of rankings across templates. Our readout contrast changes the extraction
protocol itself, from the first token to an answer line, and we ask the same question of the tool loop.

\paragraph{Written form of the input.} The written form of the input moves accuracy on basic graph tasks
\citep{fatemi2023talk}, on molecule tasks, where no text representation is best on every task
\citep{raja2026rethinking,runcie2025assessing}, and on written scores \citep{zhao2025abc,spanio2026can}.

\paragraph{Tool use and the decision to call.} ReAct interleaves reasoning and actions \citep{yao2022react}, program-aided
models execute a written program \citep{gao2022pal}, and Toolformer trains a model to decide when to call
\citep{schick2023toolformer}. Tools help a chemistry agent on structure tasks \citep{yu2024chemtoolagent} and a music
controller under one wording \citep{wang2025csymr}. The decision to call is an evaluation object of its own: When2Call scores when to call,
ask or decline \citep{ross2025when2call}, \citet{wu2026call} compare no-tool, always-tool and self-decided calls, and
\citet{cheng2026knowing} and \citet{sun2026when2tool} probe whether a model knows a call is needed before it acts.
\citet{zou2026cobra} learn the boundary from counterfactual margins. Robustness work perturbs the tools
\citep{ye2024rotbench}, their descriptions, which move tool selection by an order of magnitude \citep{faghih2025tool},
or the stages of the tool pipeline \citep{zheng2026toolrobustbench}.
Rephrasings of the user's request lower the match of generated function calls with the reference call, with much of
the loss in parameter values \citep{rabinovich2025robustness}, query paraphrases are among the perturbations that
tool-use agents withstand best in a simulated benchmark \citep{zhou2026simulation}, and indirect requests make email
agents omit actions \citep{chen2026lost}. We measure what the wording does to call frequency inside an executed loop
and to the comparison of two models, against controls that supply the report without a call.

\section{Conclusion}

We find that tools raise the accuracy of language models where a program answers the question and widen the wording
sensitivity of pairwise comparisons against reasoning without tools. In our plain-text loop, molecule accuracy follows
call frequency, and supplied reports leave the comparisons at their most stable. In molecules, native calling and one
tool description remove the detected widening in the groups tested, while other interventions have model-specific
effects. Sensitivity persists on graphs and where tools return intermediate results under the tested native
interface. We hope that evaluations of tool-using models will report accuracy, call frequency, stopping behavior and
comparison stability under several wordings.

\bibliographystyle{named}
\bibliography{references}

\clearpage
\section*{Supplementary Material}
\setcounter{section}{0}\renewcommand{\thesection}{S\arabic{section}}
\setcounter{figure}{0}\renewcommand{\thefigure}{S\arabic{figure}}
\setcounter{table}{0}\renewcommand{\thetable}{S\arabic{table}}

\section{The wordings}
\label{sup:wordings}

Each domain has five wordings of the instruction. A wording is a template. The placeholders \textit{<question>},
\textit{<input>}, \textit{<form>} and \textit{<options>} are filled with the item's question, its written input,
the name of the written form shown and the four options (letter domains), and the number domains have no options.
The same five templates serve every written form of a domain. Under tools, the tool description printed after each
domain's wordings precedes the wording unchanged. Under constant tools the description is the same and every tool
answers ``no information''. Under reasoning without tools, the sentence printed last precedes the wording in the
letter domains. The wordings differ in their output-format clause, and the variant without that clause, used in
Section~\ref{sup:clause}, ends every wording with the same sentence.

\subsection*{Symbolic music}
\noindent\textbf{p1.}\\
{\small\ttfamily \textit{<question>}\\
The score is written in \textit{<form>}.\\
~\\
\textit{<input>}\\
~\\
Options:\\
\textit{<options>}\\
~\\
Answer with the letter of the correct option only.}\par\medskip
\noindent\textbf{p2.}\\
{\small\ttfamily Read the following score, given in \textit{<form>}, and answer the question.\\
~\\
Score:\\
\textit{<input>}\\
~\\
Question: \textit{<question>}\\
\textit{<options>}\\
~\\
Reply with a single letter.}\par\medskip
\noindent\textbf{p3.}\\
{\small\ttfamily Here is a piece of music in \textit{<form>}:\\
\textit{<input>}\\
~\\
\textit{<question>} Choose one of the options below.\\
\textit{<options>}\\
~\\
Respond with only the option letter.}\par\medskip
\noindent\textbf{p4.}\\
{\small\ttfamily Question: \textit{<question>}\\
Notation: \textit{<form>}\\
~\\
\textit{<input>}\\
~\\
\textit{<options>}\\
~\\
The answer is one of A, B, C, D. Give the letter.}\par\medskip
\noindent\textbf{p5.}\\
{\small\ttfamily You are given a musical score in \textit{<form>}. \textit{<question>}\\
~\\
\textit{<input>}\\
~\\
Candidates:\\
\textit{<options>}\\
~\\
Output just the letter of the best candidate.}\par\medskip
\noindent\textbf{Tool description (before every wording under tools and constant tools).}\\
{\small\ttfamily Tools are available that read the score for you. To use one, write a line of the form CALL <name>() with <name> one of: count\_bars (the number of bars), time\_signature (the time signature written in the score, or none), key\_estimate (the key signature written in the score, if any, and the key estimated from the notes), pitch\_histogram (the six most frequent pitch names with their counts), last\_bar\_notes (the notes of the last bar). After each call you receive a line REPORT <name>: <result>. You may make at most 6 calls. When you are ready, write a line of the form ANSWER: <letter>.}\par\medskip
\subsection*{Graphs}
\noindent\textbf{g1.}\\
{\small\ttfamily \textit{<question>}\\
~\\
\textit{<input>}\\
~\\
Options:\\
\textit{<options>}\\
~\\
Answer with the letter of the correct option only.}\par\medskip
\noindent\textbf{g2.}\\
{\small\ttfamily Read the following graph description and answer the question.\\
~\\
Graph:\\
\textit{<input>}\\
~\\
Question: \textit{<question>}\\
\textit{<options>}\\
~\\
Reply with a single letter.}\par\medskip
\noindent\textbf{g3.}\\
{\small\ttfamily Here is a graph:\\
\textit{<input>}\\
~\\
\textit{<question>} Choose one of the options below.\\
\textit{<options>}\\
~\\
Respond with only the option letter.}\par\medskip
\noindent\textbf{g4.}\\
{\small\ttfamily Question: \textit{<question>}\\
~\\
\textit{<input>}\\
~\\
\textit{<options>}\\
~\\
The answer is one of A, B, C, D. Give the letter.}\par\medskip
\noindent\textbf{g5.}\\
{\small\ttfamily You are given an undirected graph. \textit{<question>}\\
~\\
\textit{<input>}\\
~\\
Candidates:\\
\textit{<options>}\\
~\\
Output just the letter of the best candidate.}\par\medskip
\noindent\textbf{Tool description (before every wording under tools and constant tools).}\\
{\small\ttfamily Tools are available that read the graph for you. To use one, write a line of the form CALL <name>(<arguments>) with one of: count\_nodes() (the number of nodes), count\_edges() (the number of edges), degree(<node>) (the number of edges at a node), neighbors(<node>) (the nodes connected to a node), has\_edge(<node>, <node>) (yes or no), nodes\_on\_cycles() (the nodes that lie on at least one cycle). Nodes are named as in the graph description. After each call you receive a line REPORT <name>: <result>. You may make at most 6 calls. When you are ready, write a line of the form ANSWER: <letter>.}\par\medskip
\subsection*{Molecules}
\noindent\textbf{y1.}\\
{\small\ttfamily \textit{<question>}\\
The molecule is written as \textit{<form>}.\\
~\\
\textit{<input>}\\
~\\
Options:\\
\textit{<options>}\\
~\\
Answer with the letter of the correct option only.}\par\medskip
\noindent\textbf{y2.}\\
{\small\ttfamily Read the following molecule, given as \textit{<form>}, and answer the question.\\
~\\
Molecule:\\
\textit{<input>}\\
~\\
Question: \textit{<question>}\\
\textit{<options>}\\
~\\
Reply with a single letter.}\par\medskip
\noindent\textbf{y3.}\\
{\small\ttfamily Here is a molecule written as \textit{<form>}:\\
\textit{<input>}\\
~\\
\textit{<question>} Choose one of the options below.\\
\textit{<options>}\\
~\\
Respond with only the option letter.}\par\medskip
\noindent\textbf{y4.}\\
{\small\ttfamily Question: \textit{<question>}\\
Notation: \textit{<form>}\\
~\\
\textit{<input>}\\
~\\
\textit{<options>}\\
~\\
The answer is one of A, B, C, D. Give the letter.}\par\medskip
\noindent\textbf{y5.}\\
{\small\ttfamily You are given a molecule as \textit{<form>}. \textit{<question>}\\
~\\
\textit{<input>}\\
~\\
Candidates:\\
\textit{<options>}\\
~\\
Output just the letter of the best candidate.}\par\medskip
\noindent\textbf{Tool description (before every wording under tools and constant tools).}\\
{\small\ttfamily Tools are available that read the molecule for you. To use one, write a line of the form CALL <name>() with <name> one of: heavy\_atom\_count (the number of non-hydrogen atoms), ring\_count (the number of rings), molecular\_weight (in g/mol), molecular\_formula, h\_bond\_donors and h\_bond\_acceptors (Lipinski's counts). After each call you receive a line REPORT <name>: <result>. You may make at most 6 calls. When you are ready, write a line of the form ANSWER: <letter>.}\par\medskip
\subsection*{Mathematics (MATH-500)}
\noindent\textbf{x1.}\\
{\small\ttfamily \textit{<question>}\\
~\\
Solve the problem. Give the final answer as a number on the last line, in the form ANSWER: <number>.}\par\medskip
\noindent\textbf{x2.}\\
{\small\ttfamily Read the following math problem and work out the answer.\\
~\\
Problem: \textit{<question>}\\
~\\
End with one line of the form ANSWER: <number>, where <number> is the final numeric answer.}\par\medskip
\noindent\textbf{x3.}\\
{\small\ttfamily Here is a math problem:\\
\textit{<question>}\\
~\\
Work through it step by step, then write the final numeric answer on its own line as ANSWER: <number>.}\par\medskip
\noindent\textbf{x4.}\\
{\small\ttfamily Question: \textit{<question>}\\
~\\
The answer is a single number. Compute it and finish with the line ANSWER: <number>.}\par\medskip
\noindent\textbf{x5.}\\
{\small\ttfamily You are given a math problem. \textit{<question>}\\
~\\
Find the number it asks for. Your last line must read ANSWER: <number> and contain only the final number.}\par\medskip
\noindent\textbf{Tool description (before every wording under tools and constant tools).}\\
{\small\ttfamily Two tools are available. CALL calculator(<expression>) evaluates an arithmetic expression over numbers with + - * / \textasciicircum{} ( ) and sqrt, pi, exp, log, sin, cos, tan. CALL solve(<equation>, <variable>) solves an equation or a system of equations separated by ; for the variable or variables named (for example solve(x\textasciicircum{}2 - 5*x + 6 = 0, x)). After each call you receive a line REPORT <name>: <result>. You may make at most 6 calls. When you are ready, write a line of the form ANSWER: <number>.}\par\medskip
\subsection*{Tables (FinQA)}
\noindent\textbf{t1.}\\
{\small\ttfamily \textit{<question>}\\
~\\
\textit{<input>}\\
~\\
Answer the question from the table. Give the final answer as a number on the last line, in the form ANSWER: <number>.}\par\medskip
\noindent\textbf{t2.}\\
{\small\ttfamily Read the following table and answer the question.\\
~\\
Table:\\
\textit{<input>}\\
~\\
Question: \textit{<question>}\\
~\\
End with one line of the form ANSWER: <number>, where <number> is the final numeric answer.}\par\medskip
\noindent\textbf{t3.}\\
{\small\ttfamily Here is a table from a financial report:\\
\textit{<input>}\\
~\\
\textit{<question>} Work through it step by step, then write the final numeric answer on its own line as ANSWER: <number>.}\par\medskip
\noindent\textbf{t4.}\\
{\small\ttfamily Question: \textit{<question>}\\
~\\
\textit{<input>}\\
~\\
The answer is a single number. Compute it and finish with the line ANSWER: <number>.}\par\medskip
\noindent\textbf{t5.}\\
{\small\ttfamily You are given a table. \textit{<question>}\\
~\\
\textit{<input>}\\
~\\
Find the number the question asks for. Your last line must read ANSWER: <number> and contain only the final number.}\par\medskip
\noindent\textbf{Tool description (before every wording under tools and constant tools).}\\
{\small\ttfamily Tools are available that read the table for you. CALL rows() lists the row labels; CALL lookup(<row label>) gives the values of one row with their column headers; CALL column(<column header>) gives the values of one column by row; CALL calculator(<expression>) evaluates an arithmetic expression over numbers and + - * / ( ). After each call you receive a line REPORT <name>: <result>. You may make at most 6 calls. When you are ready, write a line of the form ANSWER: <number>.}\par\medskip
\noindent\textbf{Reasoning without tools (letter domains; before every wording).}\\
{\small\ttfamily You may write your reasoning first. When you are ready, write a line of the form ANSWER: <letter>.}\par

\section{Items, tools, models and harness}
\label{sup:setup}

\paragraph{Items.} Molecules: 504 four-option items on compounds drawn from PubChem (2,105 of 3,000 random compounds
pass size and element filters), 84 per property: heavy-atom count, ring count, molecular weight, molecular formula, and
Lipinski hydrogen-bond donors and acceptors, with gold answers computed by RDKit. The InChI states the molecular formula,
so formula items under InChI can be answered by lookup (Section~\ref{sup:form}). Graphs: 504 four-option items following
the tasks of Fatemi et al.\ (node count, edge count, node degree, edge existence, connected nodes and cycle membership,
84 per task) on random graphs of 5 to 20 nodes from six families (Erd\H{o}s--R\'enyi, Barab\'asi--Albert, stochastic
block, star, path and complete), up to nine questions per graph. Music: the 386 items of ABC-Eval that fit a common
token budget in ABC, MusicXML and a note-event list, less 25 next-bar items whose options repeat a text in one
notation. Tables: 500 test questions of FinQA, sampled from the 692 whose supporting facts lie in the table and whose
answer is a number, shown without the report text. Mathematics: the 336 MATH-500 problems whose answer is a plain
number.

\paragraph{Scoring.} A letter answer is correct when it is the gold option, and an answer that cannot be parsed is
incorrect. A mathematics answer is correct when it equals the gold number. A table answer is correct when it lies
within 1\% of the executed answer or within 0.005 of it, or when it has at least one decimal (two when the answer is
below one) and equals the answer rounded to the decimals the model wrote. For a question whose gold value is a ratio
(its FinQA program divides, the question names a percentage, rate, margin, portion, return, growth, yield or share, or
the table lists percentages), the same rules apply to the value times 100. An integer never matches through rounding.
Under a strict rule without the percentage conversion and the tolerance, every contrast of the table rows
keeps its direction and 78\% of the Holm outcomes stay. 34 FinQA items have an executed answer that disagrees with the
answer in their text, and removing them moves the wider counts of the table rows by at most 3.

\paragraph{Inclusion criterion.} A domain enters only if, on its first 200 items under the fixed readout, at least one of
Qwen3-8B and Phi-4 scores at most 0.90 and the wording spread of the pair is at least 0.02 under one written form. GSM8K
fails it, with accuracies of 0.95 and 0.97. Music, where the protocol was developed, is exempt.

\paragraph{Tools.} Molecules: six RDKit operators (2026.03) for the six properties. Graphs: six NetworkX operators (3.7)
for the number of nodes and edges, the degree and the neighbors of a node, whether two nodes share an edge, and the nodes
on a cycle. Music: five music21 operators (10.5) for the number of bars, the written time signature, the written key
signature with the key estimated from the notes, the six most frequent pitch names, and the notes of the last bar. Run
alone, they agree with the gold option on 0.99 of bar-count items, 0.63 of key items and 0.53 of time-signature items.
Tables: three operators for the row labels, one row and one column, and a calculator. Mathematics: a calculator and an
equation solver built on SymPy 1.14. A call that names no tool, or passes an argument the tool cannot use, receives an
error line and counts as a rejected call. The tool description is one text per domain, printed before the wording.

\paragraph{Harness.} The loop generates at most 64 tokens per step in the letter domains and 256 in the number
domains, within an episode budget of 256 tokens in music and molecules, 384 on graphs, 1,024 in tables and 1,536 in
mathematics, and ends after six calls. An episode that reaches its budget without an answer line receives an appended
answer line, which is read as under the fixed readout. Under the native interface the same tools are declared through
the model's chat template, a call is a structured turn, the step budget is 128 tokens, and a model that writes prose
before its call block is parsed for the block anywhere in the turn. Under the supplied report, the report that the loop
would print for the item's question type is appended as one line, and items whose question type no operator answers
receive no line. Under the noisy supplied report, a fixed share of items, chosen by a hash so that the same items are
corrupted under every wording and form, receive the report of a wrong option. The models run on vLLM 0.31 with greedy
decoding, Qwen3 with thinking disabled through the chat template. Table~\ref{tab:sup:models} lists the revisions.

\begin{table*}[t]
\centering
\scriptsize
\setlength{\tabcolsep}{4pt}
\begin{tabular}{@{}llll@{}}
\toprule
Model & Repository & Revision (first 12) & Weights \\
\midrule
Qwen3-8B & Qwen/Qwen3-8B & b968826d9c46 & bf16 \\
Phi-4 & microsoft/phi-4 & 2db69c1c3e91 & 8-bit float \\
Llama-3.1-8B & meta-llama/Llama-3.1-8B-Instruct & 0e9e39f249a1 & bf16 \\
Gemma-3-12B & google/gemma-3-12b-it & 96b6f1eccf38 & bf16 \\
Mistral-Small-24B & mistralai/Mistral-Small-24B-Instruct-2501 & 9527884be6e5 & 8-bit float \\
Qwen3-30B-A3B & Qwen/Qwen3-30B-A3B-Instruct-2507-FP8 & 5a5a776300a4 & 8-bit float \\
Qwen2.5-72B & Qwen/Qwen2.5-72B-Instruct-GPTQ-Int8 & -- & 8-bit integer \\
\bottomrule
\end{tabular}
\caption{The models of the grid and the 72B model, with the Hugging Face revision used.}
\label{tab:sup:models}
\end{table*}

\section{Statistical procedure}
\label{sup:stats}

\begin{table}[t]
\centering
\scriptsize
\setlength{\tabcolsep}{3pt}
\begin{tabular}{@{}llccc@{}}
\toprule
Domain & Contrast & Rows & Bootstrap & Exchange \\
\midrule
Molecules & tools vs.\ fixed & 60 & 34/2 & 39/1 \\
Molecules & tools vs.\ reasoning & 60 & 37/2 & 41/1 \\
Molecules & fixed vs.\ first token & 60 & 0/5 & 0/11 \\
Graphs & tools vs.\ reasoning & 45 & 12/0 & 14/0 \\
Graphs & fixed vs.\ first token & 45 & 0/36 & 1/39 \\
Music & tools vs.\ reasoning & 45 & 17/0 & 19/0 \\
Tables & tools vs.\ reasoning & 45 & 15/0 & 14/0 \\
Tables & constant vs.\ reasoning & 45 & 32/0 & 37/0 \\
Mathematics & tools vs.\ reasoning & 15 & 5/0 & 5/0 \\
\bottomrule
\end{tabular}
\caption{Detected wider/narrower pairs under the item bootstrap and under the within-item exchange test, after Holm
correction, for the main contrasts. No pair changes direction between the two.}
\label{tab:sup:tests}
\end{table}

Intervals are 95\% percentile intervals from 1,000 bootstrap resamples over items, in which one resample drives both
models, all wordings and both conditions of a contrast. The two-sided p-value of a contrast is twice the share of
resamples on the other side of zero, floored at one over the number of resamples, and Holm correction runs across the
model pairs of one domain, written form and contrast. Under a null of equal, nonzero wording effects (one condition's
rows split at random into two pseudo-conditions within each item, three replicates), this procedure rejects at 0.05 in
0.025 of cases overall, at most 0.04 in every contrast but one, and 0.12 for tools against the fixed readout in
molecules. The within-item exchange test used for that row permutes the two conditions' outcome vectors within each
item (both models' vectors over the five wordings swap condition labels together) over 4,000 draws and rejects single
comparisons at the 0.05 level in 0.067 of 180 simulated contrasts under the same null (12 of 180; a binomial standard
error of 0.02), and in none after Holm. The simulated null splits one condition's rows into two pseudo-conditions of
equal accuracy level, so neither procedure is shown to hold the nominal rate when the two conditions differ in level,
as tools and the fixed readout do. The counts are detections and the raw differences carry the size
(Table~\ref{tab:sup:tests}). It gives 39 wider and 1 narrower pairs where the
bootstrap gives 34 and 2, and in every other contrast its counts lie within 16\% of the bootstrap's with no reversal. A
permutation test under the null of no wording effect at all is calibrated for its own null and rejects more often,
because the fixed readout has a wording effect of its own. The spread is a range, so the per-pair table also carries
the standard deviation of the difference over wordings, and every pair that is wider under the range is wider under the
standard deviation. Up to nine graph questions share one graph and up to four table questions share one table. A
cluster bootstrap over graphs and tables moves a few counts on graphs and in tables and changes no direction.

\section{Supporting results}
\label{sup:results}

\begin{figure*}[t]
\centering
\includegraphics[width=\textwidth]{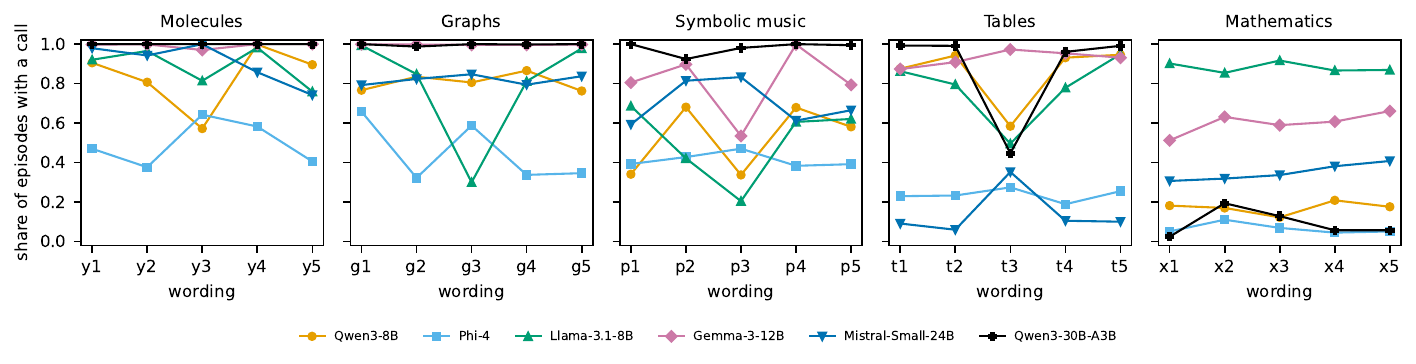}
\caption{Share of tool episodes with at least one call, by wording, one line per model and one panel per domain, under
the plain-text loop. The wordings are numbered as in Section~\ref{sup:wordings}, where the third wording of each letter
domain asks for the answer alone and the third table wording asks for step-by-step work before the answer. A flat line means a stable aggregate call frequency; it does not
establish that the same items receive calls under every wording.}
\label{fig:sup:callrate}
\end{figure*}

Counts of wider or narrower pairs are taken after Holm correction as in the main text. Figure~\ref{fig:sup:callrate}
is the call frequency of every model under every wording, referred to from the main text as Figure S1.

\subsection{The call term by comparison}

\begin{figure*}[t]
\centering
\includegraphics[width=\textwidth]{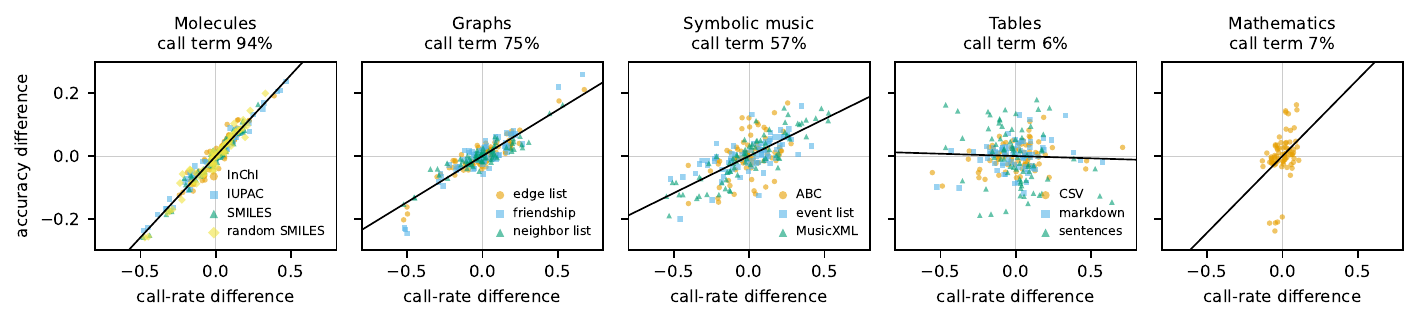}
\caption{Under tools, each point shows how far a model pair's accuracy difference and its call-rate difference move
from the pair's mean at one wording and written form. The line is the least-squares fit through the origin, and the
panel label gives the share of the variance of pair differences that the call term carries.}
\label{fig:sup:call}
\end{figure*}

Figure~\ref{fig:sup:call} shows the call term behind the call shares of the main text. The shares with their bootstrap
intervals are 0.94 [0.88, 1.02] in molecules, 0.75 [0.65, 0.81] on graphs, 0.57 [0.44, 0.62] in music, 0.06
[$-0.01$, 0.14] in tables and 0.07 [0.03, 0.11] in mathematics. The share is stable under the conventions of the split:
with the unobserved strata taking the reference value, or either order of the two factors alone, the molecule share
lies between 0.89 and 0.98, and the median over pairs is 0.93. The call rate counts whether any tool was called. The
share of item-and-form units on which a model switches between calling and not calling across wordings reaches 0.85
for Llama-3.1-8B on graphs and 0.45 for Qwen3-8B in molecules, and 0.38 for Mistral-Small-24B on graphs whose call-rate
range is 0.06, so switching exceeds the range where switches cancel.

\subsection{The forced call}

The forced call was run for Qwen3-8B, Phi-4 and Llama-3.1-8B. It raises Qwen3-8B's call rate in molecules to 0.99 or
more under every wording. The other models often answer without a call after the one refusal, so their call-rate
ranges shrink without closing: Llama-3.1-8B on graphs from 0.69 to 0.12, Phi-4 in molecules from 0.27 to 0.14. Against
tools, the forced call narrows 10 of 12 molecule comparisons, 5 of 9 graph comparisons, 3 of 9 music comparisons and no
table comparison. It widens none.

\subsection{The decision item by item}

For each model under tools we compare items whose decision to call changes across the five wordings with items whose
decision does not. In molecules, for the four models with a call-rate range above 0.1, the share of items whose
correctness also changes across wordings is 0.58 to 0.63 among the first and 0.005 to 0.28 among the second. On graphs
the ratio of the two shares is about 8 for Qwen3-8B and Mistral-Small-24B and 1.5 for Phi-4 and Llama-3.1-8B.

\subsection{Option position and permuted options}

Under the fixed readout, accuracy varies with the position of the gold letter, and permuting the options rotates each
item's gold letter while the item stays the same. Permuting moves accuracy by at most 0.04 for any model and domain and
changes the Holm outcome of none of the 150 comparisons of the letter domains, so the position effect averages out over
the balanced gold positions.

\subsection{The written form of molecules}
\label{sup:form}

Under the fixed readout, the InChI raises molecule accuracy over the PubChem SMILES by 0.06 [0.03, 0.10] for
Qwen3-8B, 0.01 [$-$0.03, 0.04] for Phi-4 and 0.05 [0.02, 0.08] for Llama-3.1-8B. The InChI states the molecular
formula, and on the formula items alone the lead is 0.27 to 0.35. Without them it is $-0.05$ to 0.05. The randomized
SMILES moves no model's accuracy by more than 0.03 from the PubChem SMILES in any condition. Under tools the IUPAC name
lowers the call rate and with it the accuracy, by 0.16, 0.09 and 0.08 for the three models. An audit found 12 InChI
strings that denote a tautomer of the SMILES. Removing those items moves the InChI lead by at most 0.005 and changes
two Holm outcomes with no reversal.

\subsection{Ten wordings and held-out wordings}

For Qwen3-8B, Phi-4 and Llama-3.1-8B on graphs and in molecules, five further wordings were written for the same
requests. With all ten, the range of a comparison under tools grows over its five-wording range by a median factor of
1.60 on graphs and 2.88 in molecules, against 1.11 and 1.12 under the fixed readout, and no contrast changes sign. On
the five further wordings alone, 2 of 3 molecule pairs reverse their order beyond 0.05 under tools and none under the
fixed readout. Leaving one of the five wordings out changes the sign of a contrast in 0.10 of all cells, most of them
rows without a Holm result, and in 36 Holm-significant rows, concentrated on the wording of each domain that lowers
call rates most. With items and wordings resampled together, few widened rows keep an interval above zero (6 of 37 in
molecules), so the spreads are statistics of this wording set.

\subsection{The tool description and its position}

\begin{table}[t]
\centering
\scriptsize
\setlength{\tabcolsep}{3pt}
\begin{tabular}{@{}lccc@{}}
\toprule
 & Qwen3 8B & Phi-4 & Llama 8B \\
\midrule
description of the main text & 0.84 (0.43) & 0.50 (0.27) & 0.89 (0.22) \\
second description & 0.97 (0.07) & 0.94 (0.06) & 1.00 (0.00) \\
third description & 0.88 (0.34) & 0.64 (0.30) & 0.98 (0.11) \\
\bottomrule
\end{tabular}
\caption{Call rate in molecules under three wordings of the tool description, with the range across the five wordings
in parentheses.}
\label{tab:sup:block}
\end{table}

Table~\ref{tab:sup:block} gives the call rate in molecules under three wordings of the tool description. Tools are
wider than the fixed readout in 10, 0 and 9 of 12 pairs under the three, and over all fifteen combinations of a wording
and a description in 12 of 12 pairs. On graphs the same three models call in 0.81, 0.96 and 0.92 (Qwen3-8B), 0.45, 0.92
and 0.77 (Phi-4) and 0.79, 0.87 and 0.87 (Llama-3.1-8B) of episodes under the three descriptions. Moving the
description into the system turn changes the largest per-wording call rate by 0.24 to 0.51 for each model in each
domain, raising Qwen3-8B's call rate in molecules to 1.00 under every wording and widening Llama-3.1-8B's range there
from 0.22 to 0.52.

\subsection{The output clause}
\label{sup:clause}

\begin{table}[t]
\centering
\scriptsize
\setlength{\tabcolsep}{3pt}
\begin{tabular}{@{}lcccc@{}}
\toprule
 & Qwen3 8B & Phi-4 & Llama 8B & Qwen2.5 72B \\
\midrule
graphs, original wordings & 0.10 & 0.34 & 0.69 & 0.99 \\
graphs, clause removed & 0.21 & 0.31 & 0.09 & 0.00 \\
molecules, original wordings & 0.43 & 0.27 & 0.22 & 0.97 \\
molecules, clause removed & 0.35 & 0.21 & 0.08 & 0.00 \\
\bottomrule
\end{tabular}
\caption{Range of the call rate across wordings under tools, with the wordings' output-format clause and with every
wording ending in the same sentence.}
\label{tab:sup:clause}
\end{table}

Table~\ref{tab:sup:clause} gives the call-rate ranges with and without the output-format clause. Without it, tools are
wider than the fixed readout in 0 of 9 graph pairs (4 with the clause) and 7 of 12 molecule pairs (10) for the three
one-card models, and the fixed readout's own spreads move by up to 0.05. Qwen2.5-72B calls in 0.61 to 1.00 of episodes
under four wordings and in 0.00 to 0.03 under the wording that asks for the bare letter, with the clause, and in 1.00
under every wording without it, where its tool spreads fall to the fixed readout's in every written form.

\subsection{The native interface by domain}

\begin{table}[t]
\centering
\scriptsize
\setlength{\tabcolsep}{3pt}
\begin{tabular}{@{}lcccccc@{}}
\toprule
 & \multicolumn{2}{c}{Qwen3 8B} & \multicolumn{2}{c}{Llama 8B} & \multicolumn{2}{c}{Mistral 24B} \\
\cmidrule(lr){2-3}\cmidrule(lr){4-5}\cmidrule(lr){6-7}
Domain & text & native & text & native & text & native \\
\midrule
Molecules & 0.43 & 0.00 & 0.22 & 0.00 & 0.26 & 0.00 \\
Graphs & 0.10 & 0.01 & 0.69 & 0.01 & 0.06 & 0.07 \\
Music & 0.34 & 0.10 & 0.48 & 0.01 & 0.24 & 0.23 \\
Tables & 0.36 & 0.01 & 0.45 & 0.00 & 0.29 & 0.63 \\
Mathematics & 0.09 & 0.60 & 0.06 & 0.00 & 0.10 & 0.12 \\
\bottomrule
\end{tabular}
\caption{Range of the call rate across wordings under the plain-text loop and under the native interface.}
\label{tab:sup:native}
\end{table}

Table~\ref{tab:sup:native} gives each model's call-rate range under both interfaces. Over the three pairs of these
models, with the same cohort's plain-text loop as the reference, tools are wider than the fixed readout in 10 of 12
molecule pairs under the text loop and 0 under the native interface (3 narrower), in 4 of 9 graph pairs and 1 (2
narrower), and in 6 of 9 music pairs and 0. In tables, against
reasoning, the counts are 6 of 9 and 4, and for the native constant control 7 and 4. Every wider native pair contains
Mistral-Small-24B. In mathematics the native loop widens 2 of 3 pairs against reasoning where the text loop widens
none, and both contain Qwen3-8B. The native interface cuts the share of Llama-3.1-8B's table episodes that end on the
call budget from 0.61 to 0.08 and its mathematics share from 0.71 to 0.20, while Qwen3-8B's native constant control
ends on the budget in 0.44 of table and 0.37 of mathematics episodes. Rejected calls stay between 0.19 and 0.38 per
call. Under the native interface, Mistral-Small-24B's accuracy in tables rises from 0.46 to 0.63 and its music
accuracy range across wordings is 0.04, close to the product of its call-rate range and the gap between the given
answer and the fixed readout. The native graph call share is 0.25 against 0.75 under the text loop.

\subsection{Operator coverage on graphs}

With only the two intermediate graph operators (a node's neighbors and whether two nodes share an edge), the three
one-card models' accuracy under tools falls from 0.97, 0.92 and 0.83 to 0.87, 0.82 and 0.61 (Qwen3-8B, Phi-4,
Llama-3.1-8B) against 0.73, 0.80 and 0.63 under the fixed readout, the pooled call share falls from 0.78 to $-0.36$,
tools are wider than the fixed readout in 0 of 9 pairs (4 with the full operators) and the constant control in 3 (1).

\subsection{Wrong reports and the noisy supplied report}

With 10, 30 and 50\% of the molecule reports replaced by a wrong option, each model's accuracy falls by about the
corrupted share times its call rate (ratios 0.57 to 1.08), its call-rate range moves by at most 0.04, and the call
share over the three pairs is 1.06, 0.65 and 0.11. The supplied report with reasoning covers the three pairs of Qwen3-8B, Phi-4 and Llama-3.1-8B over the three graph
and four molecule forms (21 rows), and the noisy supplied report the fifteen pairs of six models over the same seven
forms (105 rows). The supplied report with about a quarter of the reports wrong lowers
accuracy by 0.08 to 0.16 on graphs and 0.23 to 0.26 in molecules for the six models, is wider than the fixed readout in
0 of 105 pairs and narrower in 29, and tools are wider than it in 54 of 60 molecule and 15 of 45 graph pairs. Phi-4
overrules the wrong graph report on about half of the corrupted items.

\subsection{The supplied report with reasoning}

When the model reasons before answering with the report in its prompt, Qwen3-8B, Phi-4 and Llama-3.1-8B reach 0.88 to
0.97, their spread is wider than the fixed readout's in 0 of 21 pairs, and tools are wider than it in 11 of 12 molecule
and 7 of 9 graph pairs.

\subsection{Size}

In the Qwen3 family the call-rate range under tools is 0.37 (4B), 0.10 (8B), 0.00, 0.01 and 0.00 (14B, 30B-A3B, 32B) on
graphs, 0.01, 0.43, 0.00, 0.00 and 0.00 in molecules and 0.34, 0.34, 0.01, 0.08 and 0.04 in music. Qwen2.5-72B's is 0.99,
0.97 and 0.98. With the 72B added to the six models, tools are wider than the fixed readout in 24 of the 24 molecule,
12 of the 18 graph and 11 of the 18 music pairs that contain it.

\subsection{Sampling and thinking mode}

Under sampling at temperature 0.7 with three seeds, Qwen3-8B's ordering of the five molecule wordings by call rate is
the greedy ordering (Spearman 1.00) and its range is 0.40 against 0.43. With the thinking mode of Qwen3-8B on and a
2,048-token thinking budget, the model calls a tool in 0.12 of molecule and 0.18 of graph episodes, against 0.84 and
0.81 with it off, and the budget cuts 0.84 of its thinking under the fixed readout in molecules.

\subsection{Precision and repeatability}

Phi-4, Mistral-Small-24B and Qwen3-30B-A3B run again in bfloat16 move accuracy by at most 0.022 in 59 of 60 model,
domain and condition cells and by 0.06 in one (Qwen3-30B-A3B with tools in mathematics), and no contrast changes
direction. Repeating three full letter-domain cells changes the predicted option of 0.3\% of the rows of a fixed-readout cell
(Qwen3-8B, graphs) and of 3.8\% and 7.2\% of the rows of two tool cells (Qwen3-8B, molecules; Llama-3.1-8B, music),
and moves any pair's spread by at most 0.006, 0.012 and 0.019. Repeating Phi-4's table cell under tools changes the
predicted number of 49\% of its rows and moves any pair's spread by at most 0.036. These shares count changed
predictions. A changed prediction can leave correctness unchanged. The smallest widened table
difference is 0.076. Two runs of 200 items at batch size one agree on every answer, and each agrees with the batched
run on 98.5\% of answers. A separate run of the first-token read on the music items agrees with ours on 0.95 to 0.99
of the answers of each of five models.

\subsection{Predictions stated before the runs}

Each control's reading was written down before its rows were read. The main ones, with their outcome:
\begin{itemize}
\item Tools halve the wording spread of every pair in every domain: refuted. Tools widen it for many pairs.
\item Parsing failures account for the spread that remains under tools: refuted. The call decision does.
\item The supplied report is never wider than the fixed readout, and tools are wider than it in most molecule pairs:
  held (0 of 150 and 54 of 60).
\item The supplied report with reasoning, and with wrong reports, is never wider than the fixed readout: held.
\item The native interface shrinks the widening without removing it: refuted in molecules, where it vanishes, and held
  on graphs. Its call rate exceeds 0.9 under every wording in music: held for two of three models.
\item Under the native loop, budget stops fall below 0.10 and rejected calls below 0.01 in tables and mathematics:
  refuted for most cells.
\item A four-fold budget takes Qwen3-8B's call-rate range below 0.25: refuted (0.41).
\item Removing the output clause halves the call-rate ranges: held for Llama-3.1-8B and Qwen2.5-72B, refuted for
  Qwen3-8B and Phi-4.
\item A second tool description keeps the widening: refuted for one description and held for another.
\item Wrong reports lower the widening while the call-rate ranges stay: held.
\item Ten wordings grow the range by a factor of 1.1 to 1.5: exceeded in molecules (2.88).
\item Leaving one wording out changes a contrast's sign in at most 5\% of cells: refuted (10\%).
\item The Qwen3 decision is wording-stable from 14B: held. A 72B model's first-token mass is above 0.9 and its tool
  spreads match the fixed readout's: the first held, the second refuted.
\end{itemize}

\end{document}